# ICD 10 Based Medical Expert System Using Fuzzy Temporal Logic

*P.Chinniah, **Dr.S.Muttan
*Research Scholar, Department of ECE, CEG, Anna University, Chennai, INDIA.
**Professor, Centre for Medical Electronics, CEG, Anna University, Chennai, India

*Abstract*-Medical diagnosis process involves many levels and considerable amount of time and money are invariably spent for the first level of diagnosis usually made by the physician for all the patients every time. Hence there is a need for a computer based system which not only asks relevant questions to the patients but also aids the physician by giving a set of possible diseases from the symptoms obtained using logic at inference. In this work, an ICD10 based Medical Expert System that provides advice, information and recommendation to the physician using fuzzy temporal logic. The knowledge base used in this system consists of facts of symptoms and rules on diseases. It also provides fuzzy severity scale and weight factor for symptom and disease and can vary with respect to time. The system generates the possible disease conditions based on modified Euclidean metric using Elder's algorithm for effective clustering. The minimum similarity value is used as the decision parameter to identify a disease.

## 1.0 INTRODUCTION

There are three main relevant classes of information to be accessed by physicians when trying to reach a decision concerning a medical case namely expert's opinion, colleague's opinion and medical literature (WilliamW.Melek et al 2000). The expert opinion is necessary in medic-al decision making, since there are wide variations in clinical practices. Moreover, the growing need to assess and improve quality of health care has brought to light the possibility of developing and implementing clinical practice guidelines based on expert opinions. Even though the colleague's opinion helps in accessing information about real cases which is another important source of information, an important goal to reach when dealing with real medical cases is to have simultaneous access to the expert's opinion about the same indications of the real case being treated. The increase of the information volume in each medical field, due to the emergence of new discoveries, treatments, medicines and technologies, leads to a frequent need of consulting medical literature and in particular specialized revues and journals. Certainly, due to the huge volume of this information, a classified, targeted, access is necessary.

In the field of medicine, Imprecision and Uncertainty play a large role in the process of diagnosis of disease that has most frequently been the focus of these applications. With the increased volume of information available to physicians from new medical technologies, the process of classifying different sets of symptoms under a single name and determining the appropriate therapeutic actions become increasingly difficult. A single disease may manifest itself quite differently in different patients at different disease stages. Further, a single symptom may be indicative of several different diseases and the presence of several diseases in a single patient may disrupt the expected symptom pattern of any one of them. Although medical knowledge concerning the symptom – disease relationship constitutes one source of imprecision and uncertainty in the diagnostic process, the knowledge concerning the state of the patient constitute another.

The physician gathers knowledge about the patient from past history, physical examination, laboratory test results and other investigative procedures such as x-ray and ultrasonic. Since the knowledge provided by each of these sources carries with it varying degrees of uncertainty, here we use a fuzzy temporal logic since the state and symptoms of the patient can be known by the physician with only a limited degree of precision. In this paper, we discuss about a medical expert system in which we use fuzzy logic to identify the diseases form the symptoms which helps to develop Fuzzy rules that can be stored in the knowledge base and can be fired during further decision process. Coding of medical reports to ICD is a difficult, multilevel process, in which various kinds of errors may occur (Gergely Heja, 2002).

The Fuzzy medical expert system proposed in this paper uses ICD coding to represent data and also clustering algorithm in order to generate the most possible diseases for the given symptoms. Since, Knowledge base plays a vital role in developing a medical expert system (Hudson and Cohen, 1989), we have developed a knowledge base consisting of symptoms, diseases, question sets, fuzzy severity scale, weight factor and also ICD 10 data. This expert system provides an effective user interface which consists of pages for both doctors and patients. The doctor page displays part related symptoms and diseases with provisions for giving upper bound, lower bound values and weight factor for both selected symptom and disease. In this system, an interactive patient window is provided that generates the symptoms present in the particular part of the body when the user supplies data on body parts. For this, the whole body is divided into eight parts namely head, neck, chest, abdomen, pelvic, leg, arm and back. Each part is further subdivided into many subparts. A list of possible symptoms for each subpart is displayed by the system after user inputs are received by the system so that the user can specify the location of the symptom exactly. By selecting one or more symptoms, the associated conditions are displayed by the inference engine. After the user answers the questions, the most possible diseases are generated by excluding the least possible diseases using fuzzy temporal logic based decision making. For this purpose, we use a temporal data base that stores past and current data of patients in separate tables In order to query the temporal data base effectively, we have developed a special query language called temporal structured query language (TSQL). The remainder of this paper is organized as follows: Section 2 provides a survey of related works and compares them with the work presented in this paper. Section 3 depicts the architecture of the medical expert system proposed and implemented in this work. This section explains the various components of the expert system. Section 4 explains the implementation techniques. Section 5 shows few results obtained in this work using graphs. Section 6 gives a conclusion on this work and suggests some possible future enhancements.

## 2.0 RELATED WORKS

There are many works in the literature that explains about the design and implementation of medical expert systems. Shusaku Tsumoto(1999) had proposed a web based medical expert system in which the web server provides an interface between hospital information systems and home doctors. According to them, the recent advances in computer resources have strengthened the performance of decision making process and the implementation of knowledge base (Shusaku Tsumoto 2006) operations. Moreover, the recent advances in web technologies are used in many medical expert systems for providing efficient interface to such systems. Moreover, many such systems are put on the Internet to provide an intelligent decision support in telemedicine and are now being evaluated by regional medical home doctors. Vladimir Androuchko et al (2006) proposed an expert system called Medoctor, which is a web-based system and has a powerful engine to perform all necessary operations.

The system architecture presented by them is highly scalable, modular, and accountable and most importantly enables the incorporation of new features to be economically installed in future versions. The user interface module of that system presents a series of questions in layman's language for knowledge acquisition and also to show the top three possible diseases or conditions. However this system lacks in accuracy in decisions and also it is not following the coding of diseases as per the standards. Hence, there is a need for proposing a system with increased accuracy and standard. Viviane et al (2004) proposed a medical expert system in which platform independency plays a vital role while developing their medical expert system. Time-factor in medical diagnoses is a challenging area of research since a formalization of time-varying situations for computer-assisted diagnosis systems presents a lot of open problems for investigators (Tatiana Kiseliova, et al 2001). In this paper, we propose a medical expert system that separates rules into non-temporal and temporal components in which both components can be used by the inference engine to make predictions and decisions using fuzzy and temporal rules. The major advantages of this proposed medical expert system in comparison with the existing medical expert systems are the provision of temporal data base for storing the past and present data, a knowledge Base for inference using fuzzy temporal rules, ICD coding and a user interface for knowledge acquisition and querying.

## 3.0 SYSTEM ARCHITECTURE

The architecture of the system proposed and implemented in this research work is shown in Fig.**1.** This system consists of seven major components namely user interface, ICD coding module, Inference Engine, Temporal Information Manager, Temporal Fuzzy Decision Manager and Knowledge Base. The user interface accepts details regarding symptoms, vital signs, diseases and stores them in the knowledge base during knowledge acquisition. Moreover this user interface is used to create and manipulate a temporal data base for maintaining the patient history through the temporal information manager.

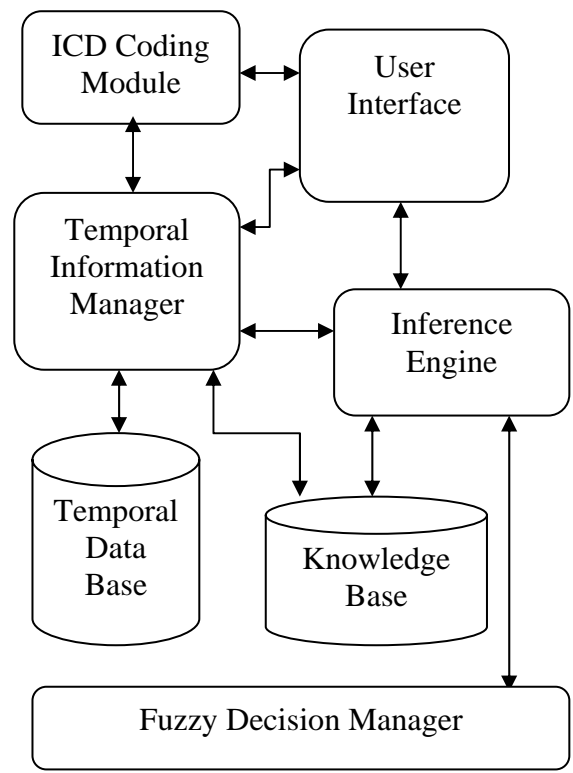

Figure1. System Architecture

The temporal information manager manages the temporal data base using TSQL commands which process queries by applying instant comparison operators and interval comparison operators. This system uses time series forecasting method to predict the future using past and present data .The temporal data base maintains two tables for each entity set to store the current data and historical data separately. The inference engine has two components namely a scheduler for scheduling the rules to be fired and an interpreter that fires the rules using forward chaining inference technique. The knowledge base is used to store rules related to patient's symptoms and diseases. The fuzzy decision manager applies fuzzy rules and coordinates with inference engine to make decision on diseases. The ICD coding module is used to store the ICD codes of diseases and symptoms, so that it is possible to make decisions using standards.

## 3.1 Temporal Database Design

The database plays a vital role in decision making since it provides the necessary facts. The temporal data base consists of symptoms, diseases, question sets from which the user can interact with system for the possible diseases (Ahamd Hansnah, 1994), fuzzy severity scale, weight factor and also ICD 10 data. There are totally 3 tables in the data base which are: (i)Tables listing all possible conditions for symptom along with ICD 10 code. (ii) Tables listing all possible symptoms for every condition along with ICD 10 code. (iii) Tables of severity scale with upper bound, lower bound values and weights for every disease and symptoms present in the human body. Here, both the severity scale and the weight vary between zero and one for any given symptom.

In this work, for the classification of diseases, the whole body is divided into eight anatomical parts and each anatomical part has many subparts. For every subpart, a symptom and disease table and a relationship between them are created. A unique code following the specification of ICD10 given by World Health Organization has been assigned to every disease and every symptom. Here, the diseases and symptoms are grouped subpart wise for better understanding and identification of diseases and symptoms in a particular part of the body. Here we have compared the number of symptoms and the number of diseases for each sub part of the body to present information about complexities that will arise if proper care is not provided to check parts like Head and Back which have large number of symptoms and diseases.

The number of symptoms and diseases has been worked out and it varies for each subpart of the human body. Table 3.1 lists the possible number of symptoms and diseases for every subpart. There are totally 839 symptoms and 4210 diseases available for all the subparts. In this work, tables are created for every subpart. Each table has symptoms and corresponding ICD 10 code with number of symptoms ranging from 10 to 80 depending on the part selected by the user. Moreover, there are 44 tables that have been created for different types of diseases. Each table has symptom and their corresponding conditions with ICD code. The total number of conditions in each table varies from 40 to 230. Along with it, there are tables for common and general symptoms and conditions. The database also contains a table for skin symptoms and diseases. For every symptom provided by the user this system generates the possible conditions using the respective tables for performing disease diagnosis.

TABLE 3.1 TABLES OF SUBPARTS

| Part | Subparts | No of symptoms | No of diseases |
|---|---|---|---|
| Head | Head | 84 | 543 |
| | Ears | 16 | 76 |
| | Eyes | 75 | 327 |
| | Nose | 20 | 101 |
| | Mouth | 66 | 248 |
| | Face | 21 | 88 |
| Neck | Neck | 38 | 221 |
| Chest | Chest | 34 | 218 |
| | Side of chest | 11 | 46 |
| | Sternum. | 16 | 94 |
| Abdomen | Upperabdomen | 22 | 166 |
| | Lowerabdomen. | 27 | 158 |
| Pelvic | Inguinal | 14 | 56 |
| | Pelvis | 23 | 83 |
| | Genital | 35 | 160 |
| | Hip | 20 | 79 |
| Arm | Fingers | 32 | 149 |
| | Palm | 23 | 102 |
| | Wrist | 11 | 66 |
| | Forearm | 16 | 56 |
| | Elbow | 20 | 89 |
| | Upper arm | 14 | 59 |
| | Shoulder | 13 | 75 |
| Leg | Foot | 21 | 94 |
| | Ankle | 13 | 83 |
| | Shin | 18 | 69 |
| | knee | 19 | 97 |
| | Thigh | 16 | 69 |
| | Toe | 18 | 97 |
| Back | Sole | 18 | 86 |
| | Calf | 17 | 75 |
| | Hamstring | 19 | 68 |
| | Back | 16 | 626 |
| | Upper spine | 14 | 64 |
| | Lower spine | 18 | 70 |

TABLE 3.2 DATABASES OF ALL SUB PARTS OF HUMAN BODY

| | | | |
|---|---|---|---|
| Ankle | Eyes Con | Inguinal Female | Pelvis Male Con |
| Ankle Con | Face | Inguinal Female Con | Register |
| Armpit | Face Con | Inguinal Male | Shin |
| Armpit Con | Fingers | Inguinal MaleCon | Shin Con |
| Back | Fingers Con | Jaw | Shoulder |
| Back Con | Foot | Jaw Con | Shoulder Con |
| Back of knee | Foot Con | Knee | Side of chest |
| Back of knee Con | Forearm | Knee Con | Side of chest Con |
| Buttock | Fore arm Con | Lower abdomen | Skin Con |
| Buttock Con | General Con | Lower abdomen Con | Sole |
| Calf | General skin symptom | Lower spine | Sole Con |
| Calf Con | General symptom | Lower spine Con | Sternum |
| Chest | Genital Female | Main parts | Sternum Con |
| Chest Con | Genital Female Con | Mouth | Thigh |
| Common Con | Genital Male | Mouth Con | Thigh Con |
| Common skin Con | Genital Male Con | Neck | Toes |
| Common skin symptom | Hamstring | Neck Con | Toes Con |
| Common symptom | Hamstring Con | Nose | Upper abdomen |
| Doctor | Head | Nose Con | Upper abdomen Con |
| Ears | Head Con | Palm | Upper arm |
| Ears Con | HipFe male | Palm Con | Upper arm Con |
| Elbow | Hip Female Con | Patient | Upper spine |
| Elbow Con | Hip Male | Pelvis Female | Upper spine Con |
| Eyes | Hip Male Con | Pelvis Female Con | Wrist |
| | | Pelvis Male | Wrist Con |

The database consisting of list of all diseases is shown in Table 3.2. The temporal data base for symptoms, signs and diseases has been created using TSQL and also by using ICD 10 code as shown in Table **3.3**. Hence the user can access any of the symptoms or signs and diseases with their corresponding ICD code for their reference.

**TABLE 3.3 SYMPTOM TABLE WITH ICD10**

| ICD 10 code | Symptom Name |
|---|---|
| R22.0 | Lump or bulge |
| R52.9 | Joint pain |
| R68.8 | Numbness or tingling |
| R68.8 | Stiffness or decreased movement |
| R68.8 | Swelling |
| R68.8 | Tenderness to touch |
| R68.8 | Warm to touch |
| R68.8 | Weakness |
| S61.0 | Bleeding |
| S63.0 | Visible deformity |
| S67.0 | Broken bone (single fracture) |

## 3.2. Knowledge Base Creation

The system has been designed in such a way that it has a strong knowledge base. The knowledge base consists of rules for 44 symptoms in which each rule has symptoms and corresponding ICD 10 code with the number of symptoms ranging from 10 to 80 depending on the part selected by the user. Moreover, there are many rules for the 44 symptoms that are created for making decisions. Each rule has symptom and their corresponding conditions with ICD code. The total number of conditions in each rule base varies from 40 to 230. Along with it, there are separate rules for common and general symptoms and conditions. It also contains a rule base for skin symptoms and diseases. For every symptom the user is selecting, the rule base generates the possible conditions from the above said table for disease diagnosis.

## 3.3 ICD10

The International Classification of Disea-ses is published by the World Health Organization (WHO). The International Statistical Classification of Diseases and Related Health Problems (most commonly known by the abbreviation ICD) provides codes to classify diseases and a wide variety of signs, symptoms, abnormal findings, complaints, social circumstances and external causes of injury or disease. Every health condition can be assigned to a unique category and given a code, up to six characters long. Such categories can include a set of similar diseases.

The ICD is used world-wide for morbidity and mortality statistics, reimbursement systems and automated decision support in medicine (William Melek and Alireza Sadeghian, 2000). This system has been designed to promote international compatibility in the collection, processing, classification, and presentation of these statistics. The ICD is a core classification of the WHO. The ICD is revised periodically by WHO and is currently in its tenth edition. The ICD is a core classification of the WHO-FIC. The ICD-10, as it is therefore known, was developed in 1992 to track mortality statistics. ICD-11 is planned for 2011 and has become the most widely used statistical classification system in the world.

### 4.0 DECISION MAKING PROCESS.

The decision making process is initiated by the inference engine whenever it receives user queries in the form of TSQL SELECT statement (Snodgrass). The TSQL SELECT statement provides an additional WHEN condition in addition to the WHERE condition provided in the SQL syntax. Using this WHEN condition, the user can ask the system to perform instant comparison, interval comparison, explanation of the past and prediction of the future. Other feature of the TSQL are automatic addition of temporal attributes with instant stamping of tuples for transaction time and interval stamping of tuples for valid time. Moreover, it provides options for history maintenance in the UPDATE and DELTE statements by extending the corresponding statements provided in SQL. In this work, decisions are made by applying the K-means clustering algorithm (Mac Queen.J.B, 1967) on the symptoms and patient histories. Then instant comparison operations such as BEFORE, AFTER and AT (Sarda.N.L, 1990)as well as Allen's interval algebra operators (James F Allen, 1983) are used for analysis of the past. For predicting the future, least square method is used in which a curve is fit based on the past and present data. From this curve, interpolation and extrapolation techniques are used to make a first level decision. A set of 5 to 10 decisions are made using temporal information manager and inference engine. They are sent to the fuzzy logic decision manager for providing weights. Based on the fuzzy scores, top 3 decisions are obtained by the inference engine. The decisions are communicated to the user interface and a user interaction is carried out. Based on this user interaction and temporal data, a final decision is derived by the inference engine.

The system developed in the present work is a user friendly and platform independent and hence it is helpful for the end user and also the medical physician as well as the knowledge engineer. On receiving the input symptoms the system generates the possible disease conditions based on modified Euclidean metric using Elder's algorithm (Elder R.C.,Esogbue A.O 1984). After login, if the patient selects the symptoms STRANGE SMELL, SNEEZING, NASAL CONGESTION, RUNNY NOSE with their severity scales as Patient X = [0.1, 0.7, 0.4, and 0.6]. With grade of membership in fuzzy set (Rudolf Seising, 2006). Then the system infers the diseases using fuzzy temporal logic by generating minimum distances as 0.39, 0.19 and 0.54 for the possible diseases namely DUST EXPOSURE, COMMON COLD and FOREIGN OBJECT IN NOSE respectively. In this case, the most likely disease corresponding to minimum value (0.19) of the distance in the similarity measure is COMMON COLD.

### 5.0 RESULTS AND DISCUSSIONS

The decision accuracy has been improved in this work as a result of introducing fuzzy temporal logic. Figure 2 shows the comparison of decision accuracy between conventional fuzzy logic based decision making and the decision making using fuzzy temporal logic. The figure 3 shows the existence of difference in the False Positive Rate by fuzzy logic method and the False Positive Rate obtained by using Fuzzy Temporal logic method. The False Negative Rate for fuzzy logic and fuzzy temporal logic is shown in Figure.3. It is clear that the decision accuracy increases with respect to increase in the number of samples and the false positive and false negative rate decrease with increase in the number of samples.

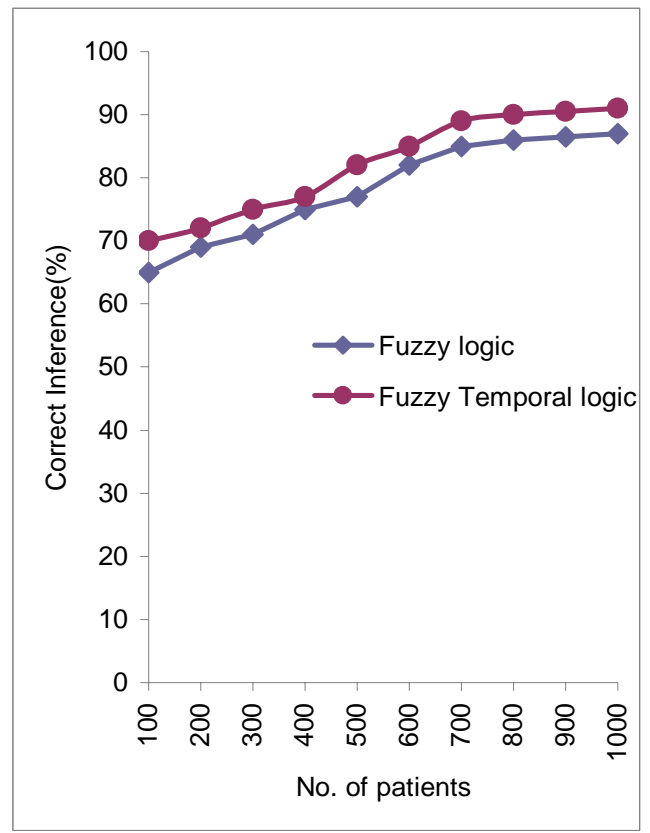

Figure 2. Accuracy of Decision making

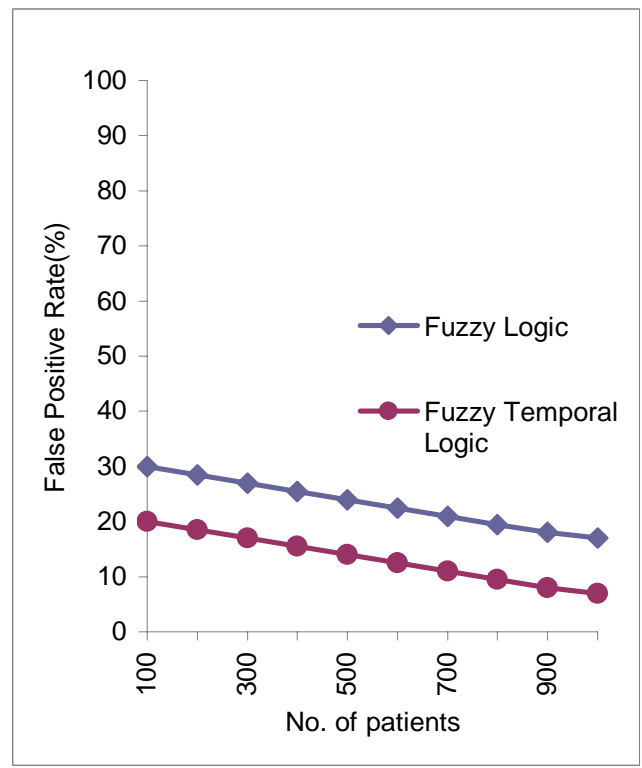

Figure 3. False Positive Analysis

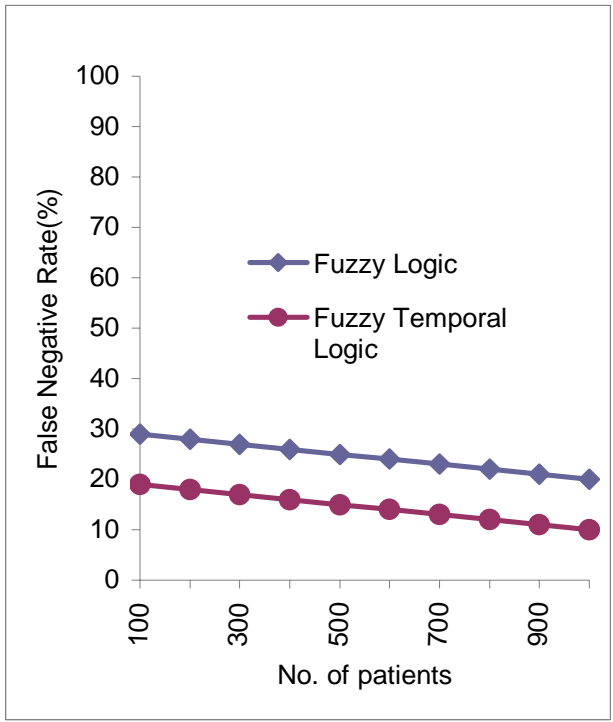

Figure 4. False Negative Analysis

## 6.0 CONCLUSION AND FUTURE ENHANCEMENTS

The expert system developed in this work is helpful to the physician, since this system not only saves time by assisting the physician but also avoids the situation where the physician needs to ask same set of questions repeatedly to the patients for the years to diagnose a disease. By this system, computerized and some intelligence added to the system in assisting the physician to take a decision. This system uses temporal logic and fuzzy logic for improving the inference process and hence inference is improved by 10% and false positive and false negative are also decreased by 7% and 9%. The major contribution of this paper is the use of ICD code, temporal rule into management and fuzzy logic based decision making. This system finds only the elementary and routine diseases in all parts of the body. In future, this system can be divided into separate expert systems that focus on only one area of human body so that the accuracy of decision making can be aided with a single domain expert

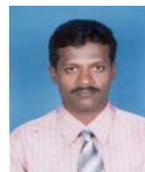

**P.Chinniah** was born in India in 1966. He received the B.Tech degree in electronics engineering and ME degree in Medical Electronics from Anna University, Chennai, India in 1992 and 1998 respectively. He is currently pursuing Ph.D. in Medical informatics Anna University, Chennai, India. He has 14 years of Teaching Experience. His research interests are in medical informatics, medical standards, e-Health and medical expert systems.

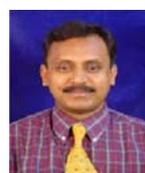

**S.Muttan** received the bachelor's degree in ECE in1985, M.E degree with honors in Applied Electronics in 1991and the Ph. D. degree in evolution and design of integrated cardiac information system in multimedia at Anna University, India in 2001. He is currently working as a professor in Centre for Medical electronics, Department of electronics and communication engineering, college of engineering Guindy, Anna University, India. His research interests are in medical informatics, pattern recognition, e-health service and biometrics.